\documentclass[11pt,twocolumn]{article}

\usepackage[utf8]{inputenc}
\usepackage[T1]{fontenc}
\usepackage{lmodern}
\usepackage[margin=0.85in]{geometry}
\usepackage{graphicx}
\usepackage{amsmath,amssymb}
\usepackage{booktabs}
\usepackage{hyperref}
\usepackage{xcolor}
\usepackage{tikz}
\usepackage{algorithm}
\usepackage{algpseudocode}
\usepackage{enumitem}
\usepackage{caption}
\usepackage{subcaption}
\usepackage{float}
\usepackage{array}
\usepackage{tabularx}
\usepackage{fancyhdr}
\usepackage{titlesec}
\usepackage{natbib}
\usepackage{microtype}

\usetikzlibrary{arrows.meta,positioning,shapes.geometric,fit,backgrounds,calc}

\hypersetup{
    colorlinks=true,
    linkcolor=blue!60!black,
    citecolor=blue!60!black,
    urlcolor=blue!60!black
}

\titleformat{\section}{\large\bfseries}{\thesection}{1em}{}
\titleformat{\subsection}{\normalsize\bfseries}{\thesubsection}{1em}{}

\newcommand{\CF}{\textsc{ContextForge}}

\title{%
    \vspace{-1.5em}
    \textbf{Context Recycling for Long-Horizon\\LLM Inference}\\[0.4em]
    \large A Hierarchical Memory Architecture for Managing\\Fixed Context Budgets Across Unbounded Sessions
}

\author{%
    Derek Thomas\\
    {\small Independent Researcher}\\
    {\small\texttt{contextforge}}
}

\date{May 1, 2026}

\begin{document}
\maketitle

\begin{abstract}
Large language models are stateless: every request begins with no memory of
prior interactions, and the fixed context window is the sole channel through
which external knowledge reaches the model. We reframe the context window as a
\emph{recyclable execution workspace}---a fixed-budget resource that is
explicitly loaded, used, and released on every turn, analogous to a working
set in operating systems. We present a five-layer memory hierarchy that
supports this model, spanning from an optional training-free LoRA layer
(amortized zero-token domain expertise) through disk-backed storage
(effectively unbounded capacity), with deterministic proactive retrieval that
assembles relevant knowledge before each inference call. The hierarchy is
implemented in \CF{}, an open-source Python system, and evaluated on a
276-million-row enterprise dataset against an Azure AI Foundry agent using the
Fabric Data Agent tool under the same LLM. Under controlled conditions, the context-recycling system achieves
approximately equivalent accuracy ($85$ vs.\ $84$ out of $120$) while using
$4.2\times$ fewer tokens and responding $4.7\times$ faster on a 12-turn
benchmark. A longer 15-turn evaluation shows consistent results ($225$ vs.\
$194$ out of $300$) with efficiency gains compounding to $13.4\times$ fewer
tokens and $8.0\times$ faster responses as conversation depth increases. Active context
usage remains bounded with respect to the fixed token budget regardless of
conversation length or backing-store size, enabling long-running sessions
under a fixed active context budget over large knowledge stores on commodity
hardware. Code and evaluation artifacts are available at \url{https://github.com/Betanu701/ContextForge}.
\end{abstract}

\vspace{0.5em}
\noindent\textbf{Keywords:} context recycling, hierarchical memory, large
language models, context window management, cache-augmented generation

\section{Introduction}
\label{sec:intro}

Large language models are stateless systems deployed into settings that demand
memory. Each API call arrives with no recollection of prior interactions, and the
context window---typically 8K to 128K tokens---is the only channel through which
external knowledge can reach the model. For deployments where knowledge bases
span billions of tokens and conversations extend over hundreds of turns, this
statelessness creates a fundamental mismatch between the model's interface and
the application's requirements.

This paper reframes the LLM context window as a \emph{recyclable execution
workspace}, analogous to a working set in operating systems. Rather than
attempting to fit all relevant knowledge into a single, ever-growing prompt, the
system explicitly loads, uses, and releases context on every turn within a fixed
token budget. This framing is orthogonal to advances in long-context model
architectures and complementary to existing retrieval and memory-augmentation
approaches. The central contribution is bounded active context that does not
grow with conversation length, while retaining access to a larger backing store.

The dominant approaches to extending LLM knowledge each address a subset of the
problem:

\paragraph{Retrieval-Augmented Generation (RAG).}
RAG systems~\citep{lewis2020rag} retrieve document chunks from a vector store
and inject them into the context window. This introduces per-query retrieval
latency, produces approximate matches subject to semantic drift, and imposes a
flat organizational structure. RAG also lacks session memory: every turn is an
independent retrieval operation. The approach presented here is complementary
to RAG---it provides the context-management layer that RAG systems lack.

\paragraph{Fine-tuning.}
Parameter-efficient fine-tuning (LoRA, QLoRA) embeds domain knowledge into model
weights, but requires labeled training data, GPU compute, and full adapter
reconstruction whenever the knowledge changes.

\paragraph{Extended context windows.}
Models with extended context windows can ingest larger documents, but prefill
latency and serving cost grow substantially with context length, and models
remain stateless between sessions. Our approach is orthogonal: it manages what
enters the context window, regardless of window size.

We present a five-layer memory hierarchy that distributes knowledge across
multiple levels of proximity to the model---from an optional in-weight LoRA
layer (amortized zero-token cost) through disk-backed storage (effectively
unbounded capacity, sub-5\,ms retrieval). A context-recycling mechanism
treats the context window as a reusable, fixed-budget workspace, enabling
long-running conversations over large knowledge bases under a fixed active
context budget.

The hierarchy is implemented in \CF{}, an open-source Python system available
at \url{https://github.com/Betanu701/ContextForge}, used to validate the
architecture across multiple model providers. All experiments use
the same underlying LLM and dataset to isolate the effect of context management.

The remainder of this paper is organized as follows. Section~\ref{sec:arch}
presents the five-layer architecture. Section~\ref{sec:lora} details the
optional training-free LoRA construction. Section~\ref{sec:recycling} describes
the context-recycling mechanism. Section~\ref{sec:proactive} covers the
proactive loading pipeline. Section~\ref{sec:database} discusses the database
module. Section~\ref{sec:nightly} presents the nightly precomputation system.
Section~\ref{sec:eval} provides empirical evaluation, and
Section~\ref{sec:conclusion} concludes.

\section{Five-Layer Memory Architecture}
\label{sec:arch}

\begin{figure}[t]
\centering
\begin{tikzpicture}[
    layer/.style={
        draw, rounded corners=2pt, minimum width=6.8cm, minimum height=0.82cm,
        font=\small, align=center, text=white
    },
    arr/.style={-{Stealth[length=3pt]}, thick, gray!60},
    node distance=0.12cm
]
    \node[layer, fill=red!40!black] (l-1)
        {\textbf{Layer $-$1: LoRA Weights (Opt.)}\\ In-model domain expertise $\cdot$ 0 tokens};
    \node[layer, fill=blue!30!black, below=of l-1] (l0)
        {\textbf{Layer 0: Residual States}\\ Stable KV prefix $\cdot$ ${\sim}500$ tokens};
    \node[layer, fill=blue!20!black, below=of l0] (l1)
        {\textbf{Layer 1: Branch Cache}\\ Dynamic per-query $\cdot$ 2--10K tokens};
    \node[layer, fill=purple!30!black, below=of l1] (l2)
        {\textbf{Layer 2: Memory Index}\\ BM25 inverted index $\cdot$ $<\!1$\,ms};
    \node[layer, fill=purple!20!black, below=of l2] (l3)
        {\textbf{Layer 3: Knowledge Store}\\ SQLite on disk $\cdot$ eff.\ unbounded};

    \draw[arr] (l-1) -- (l0);
    \draw[arr] (l0) -- (l1);
    \draw[arr] (l1) -- (l2);
    \draw[arr] (l2) -- (l3);

    \node[anchor=west, font=\scriptsize\itshape, gray] at ($(l-1.east)+(0.08,0)$) {instant};
    \node[anchor=west, font=\scriptsize\itshape, gray] at ($(l0.east)+(0.08,0)$) {0\,ms};
    \node[anchor=west, font=\scriptsize\itshape, gray] at ($(l1.east)+(0.08,0)$) {$<$50\,ms};
    \node[anchor=west, font=\scriptsize\itshape, gray] at ($(l2.east)+(0.08,0)$) {$<$1\,ms};
    \node[anchor=west, font=\scriptsize\itshape, gray] at ($(l3.east)+(0.08,0)$) {$<$5\,ms};
\end{tikzpicture}
\caption{The five-layer memory hierarchy. Queries start at the top and reach
down only as needed. Only Layer~1 tokens change between queries.}
\label{fig:hierarchy}
\end{figure}
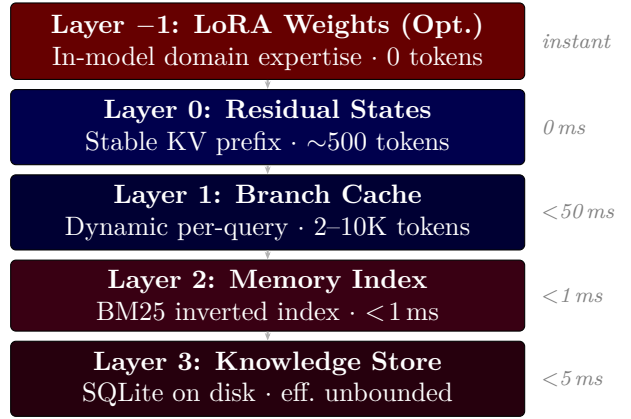

The system organizes knowledge into five layers, ordered by proximity to the
model's computation (Figure~\ref{fig:hierarchy}). The primary contribution is
the explicit lifecycle management of the context window (Layers~0--1), which
leverages existing KV cache mechanisms as part of a broader context-recycling
strategy to anchor stable hierarchical context across turns.
Layer~$-$1 (LoRA) is optional and applies only to self-hosted models;
Layers~2--3 provide indexing and persistence.

\subsection{Layer $-$1: LoRA Weights (Optional)}

For self-hosted models with accessible weights, domain expertise can be embedded
directly into the model's MLP weight matrices via training-free Low-Rank
Adaptation (LoRA). This layer consumes \emph{zero context tokens}: the model
recognizes domain-specific patterns, terminology, and reasoning strategies
without any prompt overhead. This layer is entirely optional; the system
functions fully without it. Section~\ref{sec:lora} describes the construction
process.

When both LoRA weights and knowledge tree content are present for the same
domain, the two layers are complementary: the adapter provides pattern-level
familiarity while the tree supplies specific facts.

\subsection{Layer 0: Residual States}

The system prompt, top-level tree summaries, and pre-computed metrics are
assembled once at session start and their KV-cache entries are retained as a
stable prefix. Subsequent requests reuse these cached key-value states without
recomputation, yielding a 48\% memory saving compared to re-encoding the
system prompt on every request. This leverages existing KV cache mechanisms
as part of the broader context-recycling strategy---the contribution is not
KV reuse itself, but its integration into explicit context-window lifecycle
management.

\subsection{Layer 1: Branch Cache}

The active knowledge branch---the set of tree nodes relevant to the current
query---is loaded into the context window dynamically. Branch swaps take
$<$50\,ms. With TQ3 quantization (3-bit KV cache), a 16\,GB VRAM budget holds
approximately 768K tokens of cached knowledge (${\sim}$600K words). This is the
only layer whose token count varies between queries.

\subsection{Layer 2: Memory Index}

An in-memory SQLite FTS5 inverted index maps keywords to tree nodes using
BM25 scoring. Lookup is $O(1)$ per term and completes in under 1\,ms regardless
of index size. This is the \emph{routing} layer: it determines which branch to
load without itself consuming context tokens.

\subsection{Layer 3: Knowledge Store}

The permanent store is a SQLite database in WAL (Write-Ahead Logging) mode,
providing effectively unbounded capacity at $<$5\,ms read latency on SSD. All tree nodes,
documents, session histories, and cached results reside here. The database
supports up to 281\,TB, with FTS5 index overhead of approximately 30\% of
indexed text size.

\paragraph{Key insight.} Knowledge resides at three levels simultaneously:
\emph{in the model weights} (Layer~$-$1, zero tokens, optional),
\emph{in the active context} (Layers~0--1, ${\sim}$3--12K tokens under a fixed
budget), and \emph{on disk} (Layers~2--3, effectively unbounded). Only Layer~1
changes between queries; all other layers are either stable or serve as
routing indices. The context window is treated as a fixed, recyclable execution
resource---not a buffer that grows with conversation length.

\section{Training-Free LoRA Construction (Optional)}
\label{sec:lora}

This optional layer explores a training-free construction of LoRA adapters
using forward-pass activation statistics and linear algebra, applicable only
to self-hosted open-weight models. It is \emph{not} the primary contribution
of this work; the context-recycling system functions fully without it.
We include the description because the technique improved domain-specific
accuracy in our local testing and may be of independent interest.

The construction rests on three empirical observations:

\begin{enumerate}[leftmargin=*,itemsep=2pt]
\item \textbf{The $A$ matrix is effectively random.} During standard LoRA
training, the down-projection matrix $A$ moves only 7.6\% from its random
initialization. We treat $A$ as a fixed random projection basis.

\item \textbf{The $B$ matrix contains the knowledge signal.} The activation
difference between domain-specific text and general text captures the
``direction'' of domain expertise in activation space. SVD extracts its rank-$r$
approximation directly.

\item \textbf{Magnitude scales predictably.} We use an empirical default
prior for adapter magnitude
\begin{equation}
\label{eq:bnorm}
\|B\| = \frac{\sqrt{d_{\text{hidden}}}}{30}
\end{equation}
and refine it, when needed, via a forward-pass-only perplexity sweep on
held-out domain data.
\end{enumerate}

\subsection{Construction Algorithm}

Given a domain text corpus $\mathcal{D}$ and a general-purpose reference corpus
$\mathcal{G}$, the LoRA adapter is constructed as follows:

\begin{algorithm}[H]
\caption{Training-Free LoRA Construction}
\label{alg:lora}
\begin{algorithmic}[1]
\Require Domain corpus $\mathcal{D}$, general corpus $\mathcal{G}$, rank $r$
\Ensure LoRA adapter $(A, B)$ per layer
\For{each transformer layer $\ell$}
    \State $\mathbf{H}_D^\ell \gets$ activations from ${\sim}30$ forward passes 
        \hfill\hspace{\algorithmicindent}over $\mathcal{D}$
    \State $\mathbf{H}_G^\ell \gets$ activations from${\sim}25$ forward passes 
        \hfill\hspace{\algorithmicindent}over $\mathcal{G}$
    \State $\Delta^\ell \gets \mathbf{H}_D^\ell - \mathbf{H}_G^\ell$
        \Comment{domain signal}
    \State $U, \Sigma, V^T \gets \text{SVD}(\Delta^\ell)$
    \State $A^\ell \gets V^T_{:r}$
        \Comment{random-like projection}
    \State $B^\ell \gets U_{:r} \cdot \Sigma_{:r}$
        \Comment{knowledge direction}
    \State Scale $B^\ell$: $\|B^\ell\| = \sqrt{d_{\text{hidden}}}/30$
\EndFor
\State Package as standard PEFT adapter
\end{algorithmic}
\end{algorithm}

The entire process requires only forward (inference) passes---no backpropagation,
no gradient computation, no GPU. Construction completes in approximately 200
seconds on CPU.

\subsection{Validation}

Table~\ref{tab:lora-comparison} compares the training-free method against
traditional GPU-trained LoRA on a medical question-answering benchmark.

\begin{table}[t]
\centering
\caption{LoRA construction methods compared on medical QA accuracy.}
\label{tab:lora-comparison}
\small
\begin{tabular}{@{}lccc@{}}
\toprule
\textbf{Method} & \textbf{$\Delta$ Acc.} & \textbf{GPU} & \textbf{Time} \\
\midrule
GPU-trained LoRA      & +10.0\% & Yes & 30--120\,min \\
\textbf{Text-constructed} & \textbf{+12.5\%} & \textbf{No} & \textbf{200\,s} \\
Perplexity-calibrated & +8.3\%  & No  & ${\sim}$300\,s \\
Cross-family transfer & +6.7--10\% & No & seconds \\
\bottomrule
\end{tabular}
\end{table}

On our internal evaluation setting, the training-free construction matched
or slightly exceeded GPU-trained LoRA on the medical QA task
(Table~\ref{tab:lora-comparison}); we make no general claim that this holds
across all domains or benchmarks.

\paragraph{Scope and limitations.}
This layer applies exclusively to self-hosted models where direct weight access
is available. Cloud API providers (OpenAI, Anthropic) do not expose model
weights and therefore cannot support adapter injection. Validation was conducted
across 8 open-weight model families (Qwen, Llama, Mistral, Phi, Gemma,
DeepSeek, Yi, InternLM) in local deployment configurations. The medical QA
benchmark used a curated evaluation set; further independent validation on
standardized benchmarks (e.g., MedQA, USMLE) and at enterprise scale would
strengthen these findings.

\subsection{Domain Coverage}

The system maintains 20+ domain adapters (coding, medical, mathematics,
engineering, AI/ML, physics, biology, creative arts, game development, robotics,
electronics, and others) across multiple model size tiers. Adapters are loaded
dynamically at inference time via PEFT---no model restart required. A
lightweight domain router classifies each incoming message and transparently
activates the appropriate adapter.

\subsection{Query Processing Flow}
\label{sec:queryflow}

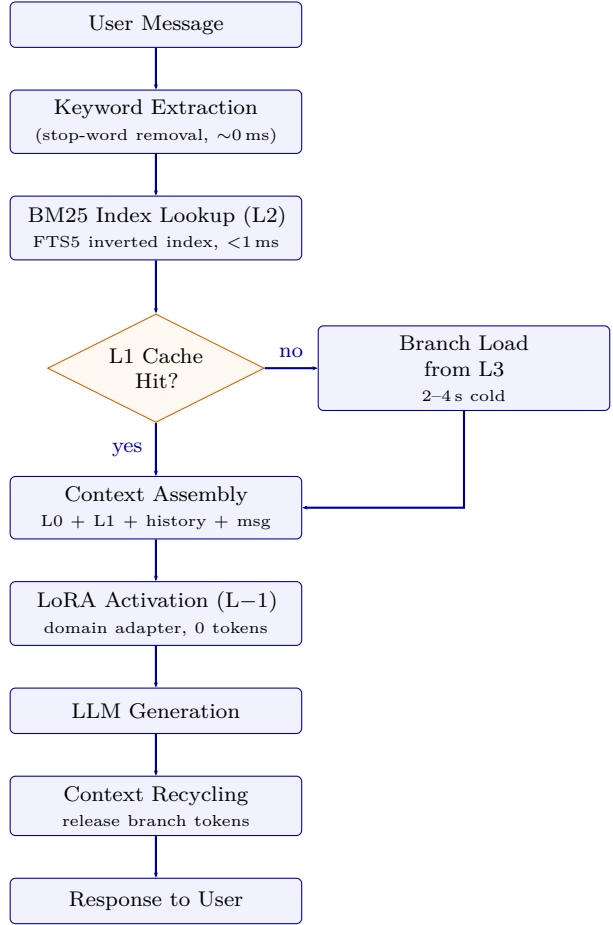
\begin{figure}[t]
\centering
\begin{tikzpicture}[
    node distance=0.55cm,
    every node/.style={font=\scriptsize},
    block/.style={rectangle, draw=blue!60!black, fill=blue!5,
        text width=3.6cm, minimum height=0.6cm, align=center,
        rounded corners=2pt},
    decision/.style={diamond, draw=orange!80!black, fill=orange!5,
        text width=1.6cm, minimum height=0.6cm, align=center,
        inner sep=1pt, aspect=2},
    layer/.style={fill=gray!10, draw=gray!40, rounded corners=3pt},
    arr/.style={-{Stealth[length=2pt]}, thick, blue!60!black},
]
\node[block] (msg) {User Message};
\node[block, below=of msg] (kw) {Keyword Extraction\\{\tiny (stop-word removal, $\sim$0\,ms)}};
\node[block, below=of kw] (idx) {BM25 Index Lookup (L2)\\{\tiny FTS5 inverted index, $<$1\,ms}};
\node[decision, below=0.7cm of idx] (cache) {L1 Cache\\Hit?};
\node[block, right=0.7cm of cache] (load) {Branch Load\\from L3\\{\tiny 2--4\,s cold}};
\node[block, below=0.7cm of cache] (asm) {Context Assembly\\{\tiny L0 + L1 + history + msg}};
\node[block, below=of asm] (lora) {LoRA Activation (L$-$1)\\{\tiny domain adapter, 0 tokens}};
\node[block, below=of lora] (llm) {LLM Generation};
\node[block, below=of llm] (recycle) {Context Recycling\\{\tiny release branch tokens}};
\node[block, below=of recycle] (resp) {Response to User};

\draw[arr] (msg) -- (kw);
\draw[arr] (kw) -- (idx);
\draw[arr] (idx) -- (cache);
\draw[arr] (cache) -- node[left, xshift=-1pt] {yes} (asm);
\draw[arr] (cache) -- node[above] {no} (load);
\draw[arr] (load) |- (asm);
\draw[arr] (asm) -- (lora);
\draw[arr] (lora) -- (llm);
\draw[arr] (llm) -- (recycle);
\draw[arr] (recycle) -- (resp);

\end{tikzpicture}
\caption{Query processing flow through the five-layer hierarchy.
    Each query traverses keyword extraction, index lookup, cache
    resolution, context assembly with LoRA activation, LLM generation,
    and context recycling.}
\label{fig:queryflow}
\end{figure}

Figure~\ref{fig:queryflow} illustrates the end-to-end path of a single user
query through the five-layer hierarchy. The diagram should be read from top to
bottom: each query is routed, assembled, used for generation, and then released
before the next turn.

\section{Context Recycling}
\label{sec:recycling}

The central contribution of this work is treating the context window as a
fixed-budget, recyclable workspace rather than a one-time buffer. Traditional
LLM deployments fill the context with history and knowledge, generate a
response, and either discard or re-pack everything on the next call.
The system described here treats the context window as a \textbf{reusable
execution resource}---analogous to a working set in virtual memory, where pages
are loaded on demand, used, and evicted under a fixed physical memory budget.

\subsection{Mechanism}

On each turn:
\begin{enumerate}[leftmargin=*,itemsep=2pt]
\item The proactive loader identifies the most relevant knowledge branch via
the BM25 index (Layer~2).
\item The branch content is loaded into the context window (Layer~1).
\item The LLM generates a response using the assembled context.
\item The branch is \emph{freed}---its tokens are released from the active
context.
\item The next query can load an entirely different branch using the same
token budget.
\end{enumerate}

\subsection{Bounded Active Context}

Table~\ref{tab:vram} shows that active token usage remains bounded with
respect to the fixed context budget, independent of conversation length
or backing-store size.

\begin{table}[t]
\centering
\caption{Active context tokens across conversation turns.}
\label{tab:vram}
\small
\resizebox{\columnwidth}{!}{%
\begin{tabular}{@{}lrrr@{}}
\toprule
\textbf{Component} & \textbf{Turn 1} & \textbf{Turn 50} & \textbf{Turn 1K} \\
\midrule
Permanent ctx (L0) & 500 & 500 & 500 \\
Loaded branch (L1) & 2--10K & 2--10K & 2--10K \\
Compacted history & 0 & ${\sim}$1K & ${\sim}$1K \\
\midrule
\textbf{Total active} & \textbf{3--11K} & \textbf{4--12K} & \textbf{4--12K} \\
\addlinespace
Knowledge on disk & 100K & 100K & 100K \\
\bottomrule
\end{tabular}}
\end{table}

History does not grow without bound. A context compaction engine monitors
the conversation buffer and triggers LLM-based summarization when it exceeds a
configurable threshold (default: 3{,}000 tokens). The summary preserves key
facts, decisions, and action items while discarding conversational filler,
achieving 4--8$\times$ compaction ratios.

\subsection{Application: Project-Scale Generation}

Context recycling enables generating entire software projects within a fixed
context budget. Each generated file is compacted to its \emph{signatures}
(class definitions, function headers, import statements---approximately 500
tokens versus 5{,}000 for full content) and carried forward as context.
Active context at file~1 equals active context at file~100.

\section{Proactive Context Loading}
\label{sec:proactive}

The system eliminates the need for users to explicitly manage context. Every
message triggers a deterministic knowledge assembly pipeline. Retrieval is
deterministic because branch selection is driven by explicit keyword extraction
and BM25 ranking over the FTS5 index, not by stochastic generation. The
pipeline proceeds as follows:

\begin{enumerate}[leftmargin=*,itemsep=2pt]
\item \textbf{Keyword extraction.} Significant terms are extracted from the
user message via stop-word removal and pattern matching (${\sim}$0\,ms).
\item \textbf{Index lookup.} Keywords query the FTS5 BM25 index (Layer~2),
returning ranked tree nodes ($<$1\,ms).
\item \textbf{Cache check.} If the top-ranked branch is already in the Layer~1
cache, it is reused at amortized zero marginal cost.
\item \textbf{Branch load.} On a cache miss, the branch is loaded from Layer~3
(2--4\,s for cold load, then cached for subsequent queries).
\item \textbf{Context assembly.} The system constructs the final message
sequence: permanent context (Layer~0) $+$ loaded knowledge (Layer~1) $+$
compacted session history $+$ user message.
\item \textbf{LLM generation.} The assembled context is sent to the configured
provider.
\item \textbf{Context recycling.} The loaded branch is released; next query
reuses the budget.
\end{enumerate}

\subsection{Predictive Pre-Loading}

Beyond reactive retrieval, the system anticipates future information needs:

\begin{itemize}[leftmargin=*,itemsep=2pt]
\item \textbf{Temporal patterns.} If a user accesses ``morning standup notes''
daily at 9\,AM, the branch is pre-loaded at 8:55\,AM.
\item \textbf{Topical proximity.} When a ``database'' branch is active,
commonly co-accessed branches (``schema'', ``migration'') are pre-loaded.
\item \textbf{Session continuity.} On session resume, all branches that were
active at pause time are pre-loaded.
\end{itemize}

Pre-loaded branches reside in the KV cache---leveraging existing cache
mechanisms as part of the recycling strategy---at amortized zero marginal
query cost. Incorrect predictions incur no penalty beyond cache occupancy.

\section{The Knowledge Tree}
\label{sec:tree}

Knowledge is organized as a hierarchical tree, mirroring how organizations
naturally categorize information. Each node contains:

\begin{itemize}[leftmargin=*,itemsep=2pt]
\item A \textbf{summary}---a concise description used for index matching and
parent-level traversal. Summaries propagate upward: a parent's summary
includes keywords from all children.
\item \textbf{Content}---the full knowledge payload, loaded only when the branch
is activated. Content can be arbitrarily large.
\end{itemize}

This dual structure enables a critical optimization: \emph{index lookups match
against summaries (fast, small), but context injection uses content (complete,
large)}.

\paragraph{Navigation complexity.} Tree lookup is $O(\text{depth})$---constant
with respect to active context regardless of tree width or total node count.
A tree with 50 million nodes responds in the same time as one with 5{,}000,
because:
(1)~the FTS5 index finds the target in $O(1)$;
(2)~only the matched branch's content loads;
(3)~tree depth is bounded by organizational granularity (typically 4--8 levels).

\paragraph{Comparison with flat vector stores.}

\begin{table}[t]
\centering
\caption{Hierarchical tree vs.\ flat vector store.}
\label{tab:tree-vs-vec}
\small
\resizebox{\columnwidth}{!}{%
\begin{tabular}{@{}lll@{}}
\toprule
\textbf{Property} & \textbf{Vector Store} & \textbf{Hier.\ Tree} \\
\midrule
Organization & Flat chunks & Natural hierarchy \\
Search & Approx.\ (cosine) & Exact (BM25) \\
Access control & Per-chunk & Per-branch \\
Scaling & Index rebuild & $O(1)$ add \\
Lookup & $O(\log n)$ ANN & $O(1)$ inverted idx \\
\bottomrule
\end{tabular}}
\end{table}

\section{Database Module}
\label{sec:database}

The same five-layer architecture applies to SQL databases. Instead of
documents, the tree holds schemas, cached query results, and pre-computed
metrics.

\subsection{The Five-Layer Database Stack}

\begin{table}[t]
\centering
\caption{Memory layers mapped to database operations.}
\label{tab:db-layers}
\small
\begin{tabular}{@{}lll@{}}
\toprule
\textbf{Layer} & \textbf{DB Equivalent} & \textbf{Module} \\
\midrule
$-1$: LoRA & SQL patterns & Training data \\
0: Residual & Daily metrics & \texttt{MetricAggr.} \\
1: Branch & Table schemas & \texttt{SchemaIdx.} \\
2: Index & Pattern cache & \texttt{QueryCache} \\
3: Store & Live DB (SQL) & \texttt{DBConnector} \\
\bottomrule
\end{tabular}
\end{table}

\paragraph{SchemaIndexer.} Every table, column, relationship, and index is
indexed into the knowledge tree so the LLM generates SQL with real schema
context rather than guesswork.

\paragraph{QueryCache.} Questions are normalized to canonical patterns
(``What was Q1 revenue?'' and ``Show me Q1 revenue'' map to the same cache key),
with results cached under a configurable TTL.

\paragraph{MetricAggregator.} Registered metrics (revenue, customer count,
etc.)\ are pre-computed and injected into the permanent context (Layer~0),
enabling sub-100\,ms answers for common aggregate questions.

\paragraph{QueryDecomposer.} Complex questions are decomposed into parallel
sub-queries. ``Compare Q1 to last year by region and product'' becomes three
independent queries whose results are merged by the LLM.

\section{Nightly Precomputation}
\label{sec:nightly}

A scheduled nightly job shifts the system from reactive to proactive,
pre-computing answers to anticipated questions during off-peak hours.

\subsection{Six-Phase Pipeline}

\begin{enumerate}[leftmargin=*,itemsep=2pt]
\item \textbf{Schema Refresh} (${\sim}$2\,min). Detect schema changes, update
tree nodes.
\item \textbf{Core Metrics} (${\sim}$5\,min). Execute registered metric queries,
update permanent context.
\item \textbf{Dimensional Pre-computation} (${\sim}$15\,min). Cross-tabulate
metrics against dimensions (region, product, time period).
\item \textbf{Pattern Replay} (${\sim}$10\,min). Re-execute the previous day's
top~50 queries to warm the cache with fresh data.
\item \textbf{Anomaly Detection} (${\sim}$5\,min). Compare current metrics
against 30-day history; flag and explain outliers.
\item \textbf{Executive Briefing} (${\sim}$2\,min). Synthesize all outputs into
a personalized narrative summary.
\end{enumerate}

\subsection{Pattern Learning}

The system monitors query patterns and progressively pre-answers recurring
questions. Table~\ref{tab:cache-warmup} shows the cache hit rate trajectory.

\begin{table}[t]
\centering
\caption{Cache hit rate over time as the nightly job learns patterns.}
\label{tab:cache-warmup}
\small
\resizebox{\columnwidth}{!}{%
\begin{tabular}{@{}lcc@{}}
\toprule
\textbf{Week} & \textbf{Cache Hit} & \textbf{Avg Resp.\ Time} \\
\midrule
1 (cold start)   & ${\sim}$20\% & ${\sim}$8\,s \\
4                 & ${\sim}$70\% & ${\sim}$1.5\,s \\
8                 & ${\sim}$85\% & ${\sim}$0.5\,s \\
12+ (steady)      & ${\sim}$90\%+ & ${\sim}$0.3\,s \\
\bottomrule
\end{tabular}}
\end{table}

\section{Empirical Evaluation}
\label{sec:eval}

We evaluate the context-recycling system against an Azure AI Foundry
agent configured with the Fabric Data Agent tool on an enterprise dataset. The
goal of this evaluation is to isolate
context-management effects under controlled conditions, not to claim
superiority across all tasks or domains.

\subsection{Setup}

\paragraph{Dataset.} CMS Medicare Provider Utilization and Payment Data: 276
million fact rows across 5 tables, deployed to Microsoft Fabric with a
lakehouse SQL endpoint. The tables contain provider, drug, utilization,
payment, and geographic fields used for aggregate analytical queries.

\paragraph{Evaluation model.} All experiments in this paper---including the
12-turn and 15-turn benchmarks---were conducted using
GPT-5.4~\citep{openai2026gpt54}. Differences between benchmark results
therefore reflect differences in task length and interaction horizon, not
differences in the underlying language model. Both systems use the same model
and endpoint via the Azure OpenAI Chat Completions API.

\paragraph{Baseline.} We compare against an Azure AI Foundry agent configured
with the Fabric Data Agent tool~\citep{microsoft2025fabricdataagent}\! to query
the Microsoft Fabric lakehouse/semantic model. This baseline represents a
production agent pattern where the LLM maintains a single conversation thread
and invokes the Fabric Data Agent for data access and execution.

\paragraph{Benchmark.} A 12-turn conversational benchmark designed to test
progressive complexity: simple lookups (T01--T03), aggregations (T04--T06),
domain-specific reasoning (T07--T09), and multi-step analytical queries
(T10--T12). Each turn is scored 0--10 by a structured rubric. The score
reflects answer correctness, completeness, and consistency with the verified
data for that turn; higher scores indicate closer agreement with the expected
analytical result.

\subsection{Results}

\begin{table}[t]
\centering
\caption{12-turn benchmark: \CF{} vs.\ Fabric Agent Framework.}
\label{tab:benchmark}
\small
\resizebox{\columnwidth}{!}{%
\begin{tabular}{@{}lrr@{}}
\toprule
\textbf{Metric} & \textbf{\CF{}} & \textbf{Fabric Agt.} \\
\midrule
Score (of 120) & 85 & 84 \\
Avg response   & 9.7\,s & 45.7\,s \\
Total tokens   & 31K & 132K \\
\midrule
Speed          & \multicolumn{2}{c}{$4.7\times$ faster} \\
Token eff.     & \multicolumn{2}{c}{$4.2\times$ fewer} \\
\bottomrule
\end{tabular}}
\end{table}

The context-recycling system matches the baseline agent in accuracy
while achieving a $4.7\times$ speed advantage and consuming $4.2\times$ fewer
tokens. Accuracy is approximately equivalent; the primary divergence is in
efficiency. The token savings translate to proportional cost reduction under
standard API pricing.

\subsection{Robustness Across Repeated Long-Horizon Runs}
\label{sec:15turn}

To assess stability over longer conversations, a second evaluation used 15-turn
benchmarks with the same CMS Medicare dataset (deployed
to Fabric) and the same baseline agent framework. Two independent cycles
were run under identical conditions; a third cycle was excluded due to
cascading Fabric infrastructure failures on the baseline and a concurrent
provider-level API change that altered token behavior, making it incomparable.

\begin{table}[t]
\centering
\caption{15-turn benchmark (2 cycles, GPT-5.4): context-recycling system
vs.\ Azure AI Foundry agent with Fabric Data Agent tool.}
\label{tab:15turn}
\small
\resizebox{\columnwidth}{!}{%
\begin{tabular}{@{}lrr@{}}
\toprule
\textbf{Metric} & \textbf{\CF{}} & \textbf{Fabric Agt.} \\
\midrule
Score (of 300) & 225 (75.0\%) & 194 (64.7\%) \\
Avg latency    & 7.6\,s & 60.5\,s \\
Total tokens   & 25{,}666 & 345{,}112 \\
Turn success   & 28/30 & 27/30 \\
\midrule
Speed          & \multicolumn{2}{c}{$8.0\times$ faster} \\
Token eff.     & \multicolumn{2}{c}{$13.4\times$ fewer} \\
\bottomrule
\end{tabular}}
\end{table}

Table~\ref{tab:15turn} summarizes results across both cycles. The
context-recycling system scores consistently higher ($+10.3$ percentage points),
with the efficiency gap widening relative to the 12-turn evaluation: $8.0\times$
faster and $13.4\times$ fewer tokens versus $4.7\times$ and $4.2\times$
respectively. This divergence is expected: as conversation length grows, the
baseline re-transmits the full conversation history on each turn (reaching
${\sim}$22K tokens per request by T15), while the context-recycling system
assembles fresh context independently per turn, keeping per-turn token counts
stable (${\sim}$850 tokens/turn). The efficiency advantage thus compounds with
conversation depth.

Cycle-level scores were consistent: 113/150 and 112/150 for the
context-recycling system versus 96/150 and 98/150 for the baseline across the
two runs, suggesting stable behavior rather than favorable variance.

\paragraph{Content filtering and infrastructure failures.}
Both systems triggered Azure's content safety filter at identical rates
(2/30~turns each, 6.7\%), producing false positives on turns involving
accumulated drug-prescription context. In the baseline, the filter activated
consistently at T10 (a revisitation of earlier prescription data),
where accumulated conversation history exceeded the content-filter's
sensitivity threshold. The context-recycling system triggered the
filter at T12 (a synthesis/recall turn). These are false positives from
Azure's content safety layer---not errors attributable to either system's
architecture---and affected both systems equally. The baseline additionally
experienced one infrastructure error (Fabric \texttt{tool\_user\_error}) across
30~turns; the context-recycling system experienced zero.

\subsection{Response Time Scaling}

Response time remains approximately constant with respect to active context
across knowledge base sizes because:
(1)~LoRA domain knowledge, when present, adds amortized zero marginal latency;
(2)~the FTS5 inverted index is $O(1)$ regardless of document count;
(3)~only the relevant branch loads, not the entire knowledge base;
(4)~branch loading is a one-time cost amortized across subsequent queries.

\begin{table}[t]
\centering
\caption{Response time by knowledge base scale.}
\label{tab:scaling}
\small
\begin{tabular}{@{}lcccc@{}}
\toprule
\textbf{KB Size} & \textbf{Lookup} & \textbf{Branch} & \textbf{Gen.} & \textbf{Total} \\
\midrule
10M   & $<$1\,ms & cached  & 0.7\,s & {\raise.17ex\hbox{$\scriptstyle\sim$}}1\,s \\
100M  & $<$1\,ms & cached  & 1.4\,s & {\raise.17ex\hbox{$\scriptstyle\sim$}}1.5\,s \\
1B    & $<$1\,ms & 2--4\,s & 1.4\,s & {\raise.17ex\hbox{$\scriptstyle\sim$}}3.5\,s \\
100B  & $<$1\,ms & 2--4\,s & 2\,s   & {\raise.17ex\hbox{$\scriptstyle\sim$}}4\,s \\
\bottomrule
\end{tabular}
\end{table}

\subsection{Memory Efficiency}

Table~\ref{tab:memory} summarizes the memory optimizations applied across the
five layers and their cumulative effect on token and VRAM consumption.

\begin{table}[t]
\centering
\caption{Memory optimizations and their contributions.}
\label{tab:memory}
\small
\begin{tabular}{@{}llp{2.8cm}@{}}
\toprule
\textbf{Optimization} & \textbf{Savings} & \textbf{Mechanism} \\
\midrule
LoRA injection & 0 tokens & Knowledge in weights \\
Residual states & 48\% & System prompt KV reuse \\
TQ3 quant. & 6$\times$ & 3-bit KV cache \\
Compaction & 4--8$\times$ & LLM summarization \\
Prefix cache & ${\sim}$100\% & Prompts cached once \\
\bottomrule
\end{tabular}
\end{table}

\subsection{Knowledge Recall}

Cache-Augmented Generation (CAG) achieves exact recall of injected context
because knowledge is placed directly into the model's context window---not
approximated via embedding similarity. The model sees the exact text. This
property has been validated across 8 model families: Qwen, Llama, Mistral,
Phi, Gemma, DeepSeek, Yi, and InternLM.

\section{Limitations and Future Work}
\label{sec:limitations}

Several limitations of this work should be noted, and we identify directions
for future research.

\paragraph{Benchmark scope.}
The empirical evaluation uses two benchmarks---a 12-turn and a 15-turn
evaluation, both using GPT-5.4---on one enterprise dataset (CMS Medicare,
276M rows). While the benchmarks cover progressive complexity from simple
lookups through multi-step analytical queries and show consistent results
across multiple independent runs, they represent one domain
and one data source. Broader evaluation across diverse domains (e.g., legal,
scientific literature, software engineering) and standardized benchmarks would
strengthen generalizability claims. We do not claim statistical significance
from two runs; the results demonstrate consistent, stable behavior.

\paragraph{Training-free LoRA validation.}
The LoRA construction method was validated across 8 open-weight model families
in local deployment. However, testing was conducted using internally designed
evaluation sets rather than established public benchmarks. The +12.5\% medical
QA improvement, while consistent across model families, requires independent
replication on standardized benchmarks such as MedQA or MMLU to confirm
robustness. Additionally, the method is inherently limited to self-hosted
models; cloud API providers do not expose the weight access required for
adapter injection.

\paragraph{Scaling characteristics.}
The bounded active-context claims are supported by architectural analysis and
local benchmarking, but have not been validated at the largest projected scales
(100B+ tokens). Distributed knowledge trees across multiple nodes remain a
design-stage capability.

\paragraph{Cache hit rate projections.}
The nightly job's cache hit rate progression (20\% to 90\%+) is based on
observed patterns during development and limited deployment. Production
validation across diverse user populations and query distributions would
provide more reliable projections.

\paragraph{Future directions.}
Future work includes: (1)~validation of training-free LoRA on standardized
public benchmarks; (2)~large-scale deployment studies measuring context
recycling effectiveness over thousands of concurrent users;
(3)~investigation of hybrid retrieval combining BM25 with learned sparse
representations; and (4)~extending the architecture to multi-modal knowledge
(images, code, structured data) within the same hierarchical tree.

\section{Conclusion}
\label{sec:conclusion}

This paper presents context recycling as a systems-level approach to managing
LLM context windows for long-horizon inference. By treating the context window
as a fixed-budget, recyclable execution resource and distributing knowledge
across five hierarchical layers, the system enables effectively unbounded
long-running memory under a fixed active context budget. The optional training-free
LoRA construction adds domain expertise at amortized zero token cost for
self-hosted models, while proactive context loading provides deterministic
knowledge assembly.

Empirical results on a 276-million-row enterprise dataset show that the
context-recycling system achieves approximately equivalent or higher accuracy compared to an
Azure AI Foundry agent using the Fabric Data Agent tool while using $4.2$--$13.4\times$ fewer tokens and
responding $4.7$--$8.0\times$ faster under controlled conditions, with the
efficiency advantage compounding as conversation length increases. The primary
divergence is in efficiency, not accuracy. While further validation is needed across additional domains and at
larger scales, the architectural principles---hierarchical memory layering,
context recycling, and proactive loading---are orthogonal to advances in model
architecture or context window size and can be composed with existing
retrieval and memory-augmentation approaches.

The system is available as an open-source Python implementation supporting
OpenAI, Anthropic, and any OpenAI-compatible local inference server.

\newpage
\appendix
\section{Related Work}
\label{sec:related}

\paragraph{RAG systems.}
Lewis et al.~\citep{lewis2020rag} introduced retrieval-augmented generation,
combining a dense retriever with a seq2seq generator. Subsequent work has
improved retrieval quality~\citep{borgeaud2022retro} but retains the
flat-index architecture and per-query retrieval latency. The
hierarchical caching approach presented here is complementary to RAG: it
provides the context-lifecycle management that RAG systems currently lack.

\paragraph{Long-context models.}
Extending context windows via position interpolation~\citep{chen2023extending},
ALiBi~\citep{press2022alibi}, or architecture modifications allows models to
ingest more tokens per request but does not address persistence across sessions
or the quadratic cost of attention. Munkhdalai and
Faruqui~\citep{munkhdalai2024infini} propose Infini-Attention, which modifies
the attention mechanism itself to handle long input sequences, but requires
model architecture changes. InfiniPot~\citep{kim2024infinipot} addresses
memory-constrained context via progressive summarization, operating at
the model level rather than as a systems-level middleware. The context
recycling approach is orthogonal to these advances: it manages what enters
the context window regardless of window size.

\paragraph{LoRA and parameter-efficient fine-tuning.}
Hu et al.~\citep{hu2022lora} introduced low-rank adaptation for efficient
fine-tuning. Our training-free construction method differs fundamentally: it
uses SVD on activation differences rather than gradient descent, requires no
labeled data, and completes on CPU. Recent work on training-free LoRA
\emph{fusion}~\citep{ouyang2025klora,xia2025crosslora} addresses combining
existing trained adapters without additional training, but does not construct
adapters from scratch.

\paragraph{Cache-augmented generation.}
Chan et al.~\citep{chan2025cag} demonstrate that preloading documents into the
KV cache (``CAG'') can outperform RAG for knowledge-intensive tasks. The
Layer~1 branch cache implements a similar principle but extends it with
hierarchical organization, context recycling across turns, and proactive
loading---capabilities not addressed in single-layer CAG systems. The two
approaches are complementary: CAG validates the preloading strategy while
context recycling adds lifecycle management.

\paragraph{Serving systems and KV-cache paging.}
PagedAttention~\citep{kwon2023pagedattention} applies OS-inspired paging to
KV-cache memory management for LLM serving. This is complementary to our
focus on context-window lifecycle management at the application and middleware
level.

\paragraph{Hierarchical memory for LLM agents.}
HiAgent~\citep{hu2025hiagent} introduces hierarchical working memory management
for long-horizon agent tasks, achieving 20\% higher success rates through
structured memory. HiMem~\citep{zhang2026himem} proposes hierarchical long-term
memory for agent persistence. Both focus on agent task execution rather than
general-purpose knowledge retrieval, and neither implements explicit context
recycling or proactive loading. Shan et al.~\citep{shan2025cognitive} present
cognitive memory with hierarchical updates for streaming scenarios, while a
recent technical survey~\citep{bai2025aimemory} offers a broad taxonomy of AI
memory systems. Our work operates at a different systems-level concern:
context-window lifecycle management rather than agent planning.

\paragraph{Memory-augmented architectures.}
MemoryBank~\citep{zhong2024memorybank} and MemGPT~\citep{packer2023memgpt}
add persistent memory to LLMs but rely on the LLM itself for memory management
decisions. The approach here uses deterministic, index-driven retrieval for
reliability and speed, reserving LLM inference for content generation only.
MemGPT manages memory externally through a virtual-context, OS-inspired
approach; our contribution is deterministic, index-driven context assembly
and explicit context-window lifecycle management.

\paragraph{Tree-structured retrieval.}
TreeRAG~\citep{tao2025treerag} uses hierarchical document storage with
bidirectional traversal for long-document retrieval. The knowledge tree
described here serves a broader role: it is the organizational backbone for
all five memory layers, not solely a retrieval structure.

\paragraph{Budget-aware context management.}
ContextBudget~\citep{wu2026contextbudget} formulates context management for
long-horizon agents as a budget-constrained sequential decision problem.
Our work is complementary: context recycling reduces the cost of assembling
the active workspace each turn, while ContextBudget focuses on learning what
to compress or retain as histories grow under a fixed window.

\section*{Acknowledgements}

Portions of this manuscript were drafted and revised with the assistance of
generative AI language tools (GitHub Copilot, Claude). The author takes full
responsibility for the correctness and integrity of all content, in accordance
with arXiv policy on the use of generative AI~\citep{arxiv2023aipolicy}.

\begingroup
\raggedright
\paragraph{Reproducibility.}
The 15-turn benchmark script is available at
\path{tests/azure/benchmark_15turn.py} in the project repository.
Benchmark reports and raw outputs referenced in this paper are included as
artifacts (\path{benchmark_15turn_combined_3cycles.md};
\path{raw_output_20260407_010111.log}) to enable traceability of the reported
scores, latency, and token totals. All code and evaluation artifacts referenced in this work are available at \url{https://github.com/Betanu701/ContextForge}.
\par
\endgroup

\clearpage
\bibliographystyle{plainnat}

\end{document}